\documentclass[runningheads]{llncs}
\usepackage[T1]{fontenc}
\usepackage{graphicx,verbatim}
\usepackage{xcolor}
\usepackage{float}
\begin{document}

\title{From Manuals to Maintenance: Fine-Tuning MedGemma for Multi-Modal Imaging System Support in Low-Resource Settings}

\titlerunning{Fine-Tuning MedGemma for Medical Device Maintenance} 
%

\author{Bernes Lorier Atabonfack\inst{1}\and
Zion Kongbi Nfo \inst{3, *}\and
Ahmed Tahiru Issah\inst{1,*} \and
Tolulope Olusuyi\inst{2, *} \and 
Clemence Ingabire \inst{1,\dag}\and 
Mohammed Hardi Abdul Baaki\inst{1,\dag}\and 
Mawuli Deku \inst{1,\dag}\and
Abdulrazaq Zubair \inst{4,\S}\and
Alyasaa Anas \inst{5,\S}\and
Raymond Confidence \inst{6}\and
Maruf Adewole\inst{2,7}  \and 
Udunna C. Anazodo\inst{2,8}\and 
} 
\authorrunning{Bernes et al.}

\institute{Carnegie Mellon University Africa, Kigali, Rwanda \and
Medical Artificial Intelligence Laboratory, Lagos, Nigeria \and 
IngenziAI, Kigali, Rwanda \and
Federal University of Health Sciences Azare, Nigeria. \and
University of Maiduguri Teaching Hospital,  Borno State, Nigeria \and
Lawson Health Research Institute, London, Canada \and
University of Pennsylvania \and
McGill University, Montréal, Canada \\
\email{
batabonf@alumni.cmu.edu,
kongbizion3@gmail.com,
aissah@alumni.cmu.edu,
tolusuyi@mailab.io,
ingabireclec@gmail.com,
abdulbaakimohammedhardi@gmail.com,
kofi.mawulideku@gmail.com,
abdulrazaq.zubair@nkdc.ng,
alyasaa390@gmail.com,
raymondconfidence@gmail.com,
madewole@mailab.io,
udunna.anazodo@mcgill.ca }}  

\maketitle   

\footnote{* Co-second authors; $\dagger$ Co-third authors; $\S$ Co-fourth authors.}
\let\thefootnote\relax\footnote{Corresponding author: Udunna C. Anazodo, udunna.anazodo@mcgill.ca}

\begin{abstract}

Imaging device downtime is a major barrier to healthcare delivery in low- and middle-income countries (LMICs), often driven by limited access to specialized biomedical engineering support. We present a multi-modality medical equipment maintenance question-answering (QA) framework and demonstrate the fine-tuning of a medical foundation model for specialized technical troubleshooting tasks. Guided by a multi-country survey across nine LMICs, we curated technical manuals from MRI and ultrasound systems to generate the INGENZI\_DatasetV1, containing 10,294 high-quality, filtered QA-context pairs. Using QLoRA-based parameter-efficient fine-tuning, we adapted the MedGemma-4b-it model to interpret system error logs and generate step-by-step equipment repair instructions. Compared to the baseline model, the fine-tuned system achieved substantial improvements across metrics, including F1 score (0.22 to 0.38), ROUGE-2 (0.18 to 0.41), and BERTScore F1 (0.86 to 0.91). These metric gains demonstrate that the model generates significantly more precise and procedurally accurate technical responses to new troubleshooting queries. This work establishes a reliable foundation for AI-assisted diagnostic and maintenance tools in resource-constrained settings.


\end{abstract}
\section{Introduction}

Medical imaging technologies such as magnetic resonance imaging (MRI), computed tomography (CT), ultrasound, and X-ray systems are essential to modern healthcare, enabling diagnosis, treatment planning, and disease monitoring across a wide range of conditions. Global demand for imaging continues to grow \cite{Equi_maintenance} \cite{who_imaging}, yet access and reliability remain highly uneven, particularly in low- and middle-income countries (LMICs), where equipment downtime significantly limits diagnostic capacity \cite{who_imaging}\cite{ref_lancet_glob}. In resource-constrained settings, prolonged downtime is often driven by limited availability of trained biomedical engineers, restricted access to service documentation, costly manufacturer-dependent servicing models, and infrastructure challenges such as unstable power and connectivity \cite{ref_bernes_et_al}. These factors reduce system availability and negatively impact patient care.

Recent advances in large language models (LLMs) and medical foundation models demonstrate strong clinical reasoning \cite{ref_medgemma}\cite{ref_medpalm} \cite{ref_llava}. However, their potential to assist biomedical technicians with troubleshooting and maintenance remains largely unexplored. Prior work introduced INGENZI Tech, an AI-powered diagnostic support for medical device troubleshooting, using general-purpose LLMs \cite{ref_bernes_et_al}. This initial framework lacked domain-specific device manual datasets for training and benchmarking AI models, limiting real-world technical precision. To address this, we present an improved framework for adapting medical foundation models to support imaging equipment. Guided by a multi-country survey across nine LMICs, we characterized local maintenance challenges and use this information to curate and publicly release the INGENZI\_DatasetV1 \cite{imaging_equipment_qa_2026}, as a corpus of 10,294 high-quality question-answer pairs derived from Siemens MAGNETOM MRI and Philips HDI 5000 ultrasound operator and service manuals. Using this dataset, we fine-tuned MedGemma-4B\cite{ref_medgemma} via Quantized Low-Rank Adaptation (QLoRA) \cite{ref_qlora}, demonstrating substantial improvements in answering operational and maintenance-related queries.

Our contributions are threefold: (1) a multi-country survey analyzing imaging equipment maintenance challenges in LMICs; (2)  the public release of the INGENZI\_DatasetV1, a multi-modality technical question-answer (QA) dataset designed for equipment support modeling; and (3) parameter-efficient fine-tuning of a medical foundation model for biomedical imaging equipment troubleshooting.

\section{Related Work}

\subsection{Medical Foundation Models}

Medical foundation models are large language models adapted to biomedical and clinical domains through domain-specific pretraining and instruction tuning. Prior work has shown that such models can achieve strong performance on medical question answering and clinical reasoning tasks. For instance, Med-PaLM \cite{ref_medpalm} demonstrated near-expert performance on standardized medical examinations when appropriately fine-tuned and aligned \cite{ref_singhal}. MedGemma follows this paradigm by providing instruction-tuned medical language models optimized for healthcare applications. While these models incorporate biomedical terminology and domain knowledge, their evaluation has largely focused on clinical reasoning benchmarks. Their use for operational support of medical equipment, such as interpreting system errors or guiding maintenance procedures, remains underexplored. In this work, we adapt MedGemma-4B to MRI system operation and maintenance using a curated dataset derived from manufacturer documentation.

\subsection{Parameter-Efficient Fine-Tuning of Large Models}

Full fine-tuning of large language models is computationally expensive. Parameter-efficient fine-tuning (PEFT) \cite{ref_peft} mitigates this by training a small set of additional parameters while freezing the base model weights. Low-Rank Adaptation (LoRA) \cite{ref_lora} introduces trainable low-rank matrices into attention layers, significantly reducing the number of updated parameters. Quantized LoRA (QLoRA) \cite{ref_qlora} further enhances efficiency by combining 4-bit weight quantization with LoRA adapters, enabling multi-billion-parameter models to be fine-tuned on modest hardware without substantial performance degradation \cite{ref_dettmers}. We employ QLoRA to adapt MedGemma-4B to the INGENZI\_DatasetV1, enabling efficient domain specialization while preserving the general medical reasoning capabilities of the base model.

\section{Method}

\subsection{Multi-Country Survey of Imaging Equipment Maintenance Practices}

To ground the development of our device operation dataset and model real-world operational challenges, we conducted a cross-sectional survey targeting biomedical engineers, radiographers, and imaging technicians working in LMICs. The survey characterized device availability, breakdown patterns, maintenance practices, access to documentation, and digital readiness across diverse healthcare settings. 

The survey was administered using Google Forms and structured into five sections covering facility characteristics, device inventory, maintenance practices, fault management, and digital readiness. Participation was voluntary. Responses were anonymized unless participants explicitly opted into follow-up testing. Consent to participate and the privacy statement indicated that responses would remain confidential and would be used solely for platform development and improvement purposes, and participants could withdraw at any time without penalty. A total of 61 facilities across nine countries (Nigeria, Rwanda, Ghana, Ethiopia, Tanzania, Uganda, Cameroon, Kenya, and Nepal) participated. The survey findings (see results section) directly informed our dataset scope and modeling pipeline. This includes common reports of high breakdown rates, prolonged downtime, limited access to documentation, and fragmented logging practices. These collectively highlight the urgent need for structured, multimodal technical knowledge on device maintenance, accessible via natural language interfaces. Based on the findings, we curated common device operator and service manuals into the multi-modal INGENZI\_DatasetV1 and used for parameter-efficient fine-tuning of a medical foundation model (MedGemma-4B) tailored specifically to imaging equipment maintenance tasks.

\subsection{INGENZI  Dataset Curation}

\subsubsection{Data Sources and Scope}
 To construct a domain-specific dataset for imaging equipment support, we curated technical documentation from two imaging modalities: Siemens MAGNETOM MRI systems and Philips HDI 5000 ultrasound systems. All technical manuals were retrieved from open-access online repositories and public engineering platforms. The MRI corpus included 19 operator, service, and maintenance manuals totaling 2,077 pages, covering multiple MAGNETOM models and software versions. The ultrasound corpus included 10 documents totaling 2,322 pages, spanning service manuals, software instructions for use, technical specifications, and cleaning/disinfection guidance. All source PDFs were used strictly for dataset construction and were not redistributed. The released INGENZI\_DatasetV1 contains only structured question–answer (QA) pairs extracted from documents, with associated textual context, and excludes proprietary images, diagrams, or raw manuals.

\subsubsection{Data processing}
\begin{figure}[h]
    \centering
    \includegraphics[width=1.0\linewidth]{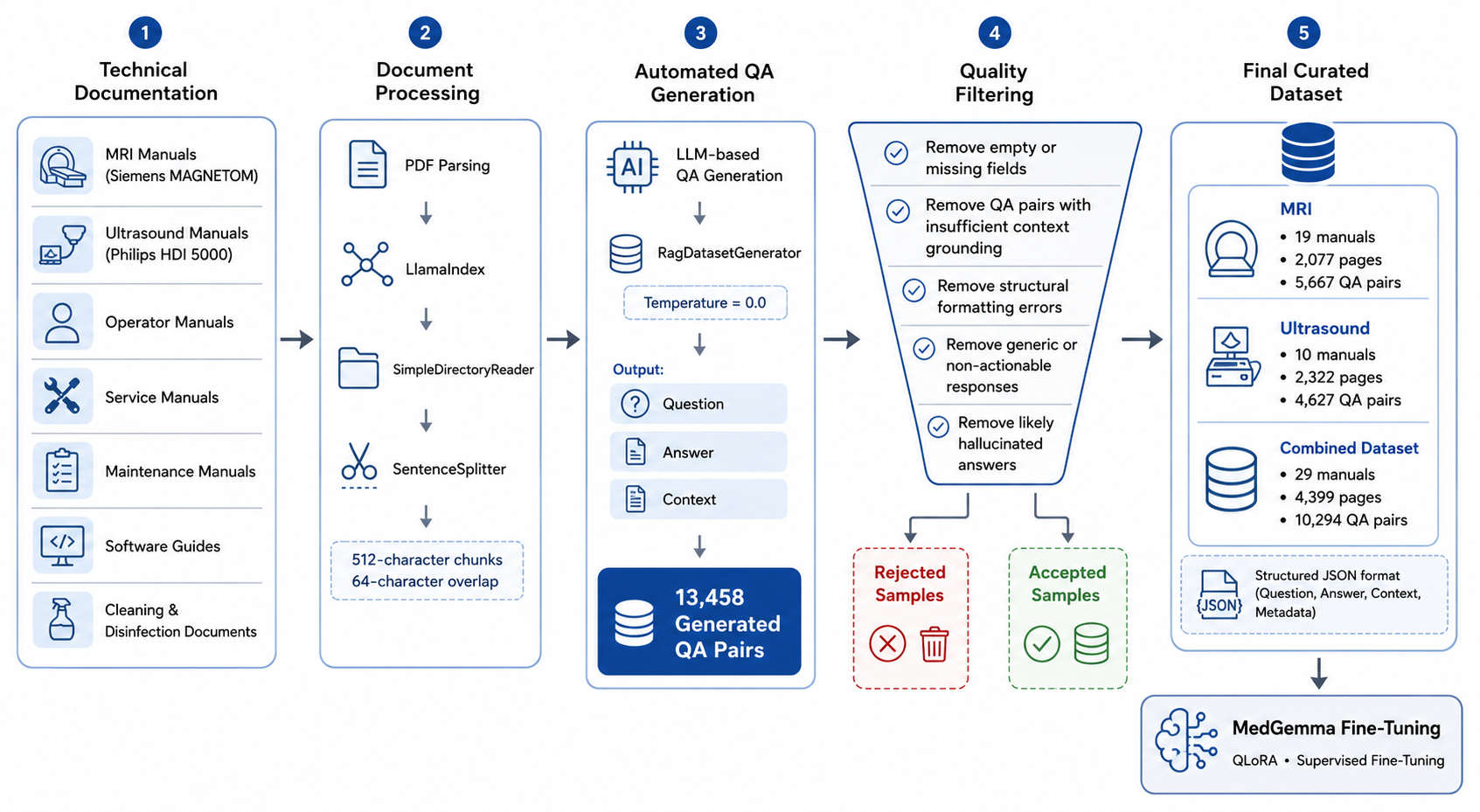}
    \caption{Workflow for constructing the medical imaging equipment QA dataset from technical manuals.}
    \label{fig:dataflow}
\end{figure}

We implemented a unified pipeline using LlamaIndex to convert unstructured technical documentation into synthetic QA pairs for supervised fine-tuning. Documents were ingested and segmented into overlapping text chunks (512 characters with a 64-character overlap) to promote the generation of grounded, specific questions. Using RagDatasetGenerator, we generated question, answer, and reference context triples. Generation temperature was set to 0.0 to reduce hallucinations, and retry mechanisms were used for robust large-scale processing. The generated samples underwent manual and rule-based filtering to remove entries with missing fields, insufficient context grounding, structural formatting errors, generic responses (e.g., “refer to the manual”), and likely hallucinated answers. For MRI documentation, 8,222 samples were initially generated, of which 5,667 high-quality QA pairs were retained. For Philips HDI 5000 ultrasound documentation, 5,236 samples were generated, with 609 removed during filtering, leaving 4,627 QA pairs. An overview of the data curation and processing is summarized in Fig \ref{fig:dataflow} 

\subsection{Model Fine-Tuning and Experimental Setup}

Using the INGENZI dataset, we fine-tuned the MedGemma-4b-it model for technical question answering using parameter-efficient fine-tuning. The task formulation was causal language modeling, where each training example was formatted into an instruction-style prompt containing the question and target answer. The INGENZI dataset was split into training, validation, and test partitions containing 8,235, 1,029, and 1,030 samples, respectively, with a random seed of 42.

\paragraph{Fine-Tuning Method.}
We employed Quantized Low-Rank Adaptation (QLoRA) to enable efficient fine-tuning under limited computational resources. The base model was loaded in 4-bit precision using BitsAndBytes quantization, and LoRA adapters were applied to all linear layers. The LoRA configuration used rank $r=16$, scaling factor $\alpha=16$, and dropout 0.05. Bias parameters were not adapted. Supervised fine-tuning (SFT) was performed using the Transformers Reinforcement Learning (TRL) library's SFTTrainer.

\paragraph{Training Configuration.}
Training was conducted on Google Colab using NVIDIA A100 Tensor Core GPUs. The model was trained for 15 epochs with a learning rate of $2\times10^{-4}$ using the fused AdamW optimizer. Batch size was 4 per device with gradient accumulation steps of 4, resulting in an effective batch size of 16. Gradient checkpointing was enabled to reduce memory usage. A linear learning rate scheduler with a warmup ratio of 0.03 was applied. The maximum gradient norm was clipped at 0.3. Models were evaluated every 50 steps, and checkpoints were saved at each epoch.

\paragraph{Evaluation Metrics.}
Model performance was evaluated using token-level F1 score, ROUGE-1\cite{rouge}, ROUGE-2, ROUGE-L, and BERTScore F1 \cite{bert_score}. The baseline corresponds to the original MedGemma-4b-it model without domain-specific fine-tuning.

\section{Results}

\begin{figure}
    \centering
    \includegraphics[width=0.9\linewidth]{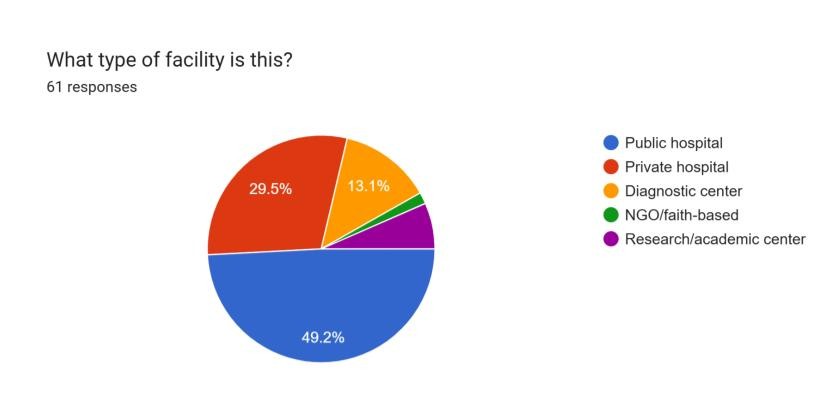}
    \caption{Survey Responses by healthcare facility type (n = 61).}
    \label{fig:facilities}
\end{figure}

\begin{figure}[h]
    \centering
    \includegraphics[width=1.0\linewidth]{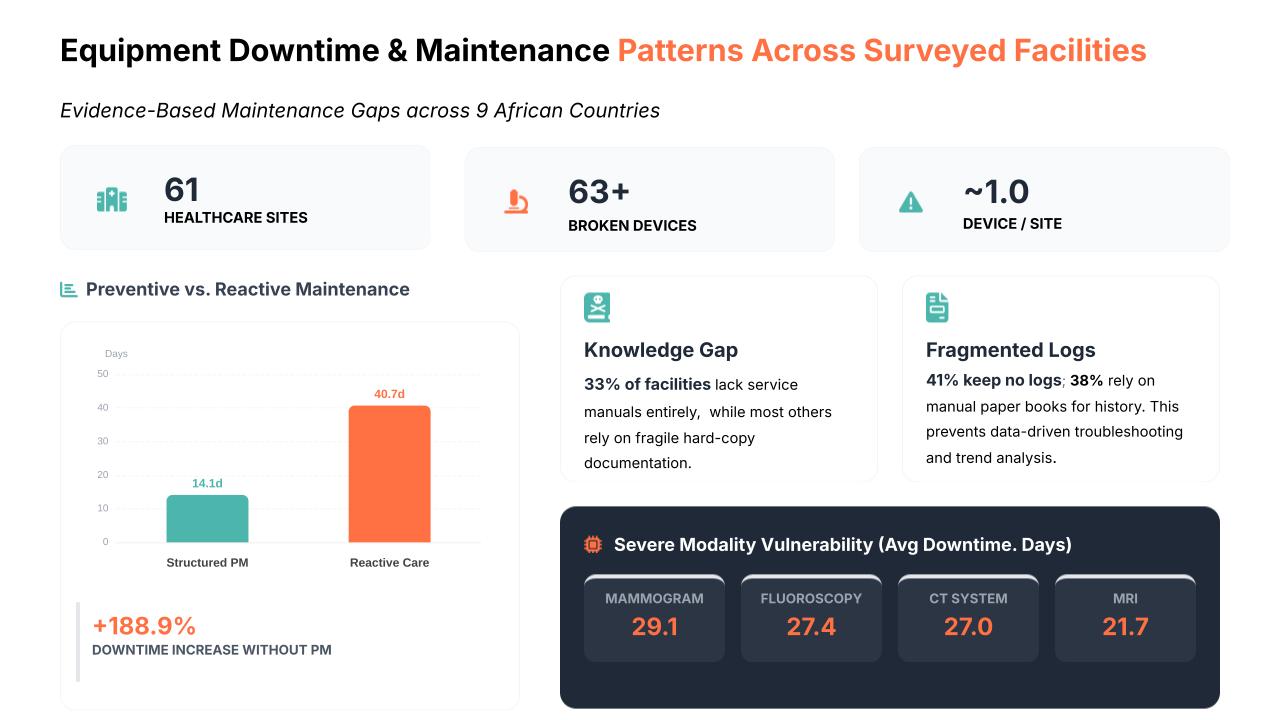}
    \caption{Summary of survey findings across 61 healthcare facilities in nine LMICs. A total of 63 non-functional imaging devices were reported, corresponding to approximately one broken device per affected site. Facilities with structured preventive maintenance exhibited substantially lower mean downtime (14.1 days) compared to reactive maintenance settings (40.7 days), representing a 188.9\% increase in downtime without structured maintenance. Mammography, fluoroscopy, CT, and MRI systems demonstrated the longest average downtime durations. Documentation gaps and fragmented logging practices were common across facilities.}
    \label{fig:survey_sum}
\end{figure}

\textbf{Survey Facility Characteristics and Device Distribution:} Among participating facilities, public hospitals constituted the largest group (30/61), followed by private hospitals (18), diagnostic centers (8), research and/or academic centers (4), and NGO/faith-based institutions (1) Fig \ref{fig:facilities}. Across facilities, ultrasound and X-ray systems were the most widely available modalities, whereas SPECT systems were relatively rare. Breakdown frequency was high across commonly used modalities. In total, 63 devices were reported as non-functional across the 61 facilities, corresponding to an average of more than one broken imaging device per facility Fig \ref{fig:survey_sum}.

In terms of documentation, access to technical documentation was limited, with 41\% facilities relying exclusively on hard-copy manuals, 11\% on digital manuals, while 33\% reported no manuals available, and 15\% had both hard-copy and digital formats. Despite these limitations, 72\% (44/61) expressed willingness to share technical documentation for knowledge-building purposes. Error logging practices were inconsistent, with most facilities reporting either no formal logging system or reliance on paper-based logs, while only a minority used specialized digital systems. Open-ended responses revealed recurring themes, summarized as: 1) software-related failures (calibration errors, system freezes, PACS connectivity issues); 2) electrical instability and power fluctuations; 3) hardware failures (coil failures, tube rotor errors); 4) human resource shortages; financial constraints and spare part limitations; 5) maintenance culture deficiencies; and 6) operational overload. These findings indicate partial digital readiness, suggesting that AI-based support systems must accommodate heterogeneous connectivity environments.

\textbf{The INGENZI Dataset V1:} 
The combined corpus contains 10,294 QA pairs spanning MRI and ultrasound equipment maintenance, operation, safety, and troubleshooting tasks. Each entry contains an identifier, question, answer, supporting context, and generation metadata Table \ref{tab:dataset_schema}. No raw manuals or proprietary assets are redistributed. To our knowledge, this is among the first openly available datasets derived from medical imaging equipment documentation specifically designed for training and evaluating language models for technical maintenance support. The dataset is publicly available and can be accessed here \cite{imaging_equipment_qa_2026}

\begin{table}[h!]
\centering
\caption{Structure of each entry in the
\textit{INGENZI\_DatasetV1}}
\label{tab:dataset_schema}
\begin{tabular}{l l}
\hline
\textbf{Field} & \textbf{Description} \\
\hline
\texttt{id} & Unique identifier for the QA pair \\
\texttt{question} & Natural language technical question \\
\texttt{answer} & Grounded reference answer derived from context \\
\texttt{context} & Source text chunk from MRI documentation \\
\texttt{num\_contexts} & Number of context segments used (typically 1) \\
\texttt{query\_by} & Source of question generation (AI-generated) \\
\texttt{answer\_by} & Source of answer generation (AI-generated) \\
\hline
\end{tabular}
\end{table}

\textbf{Fine-Tuned Foundational Model:} 
Table \ref{tab:results} presents the performance comparison between the base MedGemma model and the fine-tuned model on the MRI technical QA task. Fine-tuning resulted in substantial improvements across all lexical overlap and semantic metrics. Notably, ROUGE-2 increased from 0.18 to 0.42, indicating a much stronger capture of bigram structure and procedural phrasing. Token-level F1 improved from 0.22 to 0.38 while BERTScore F1 increased from 0.86 to 0.92, demonstrating that the fine-tuned model generates significantly more accurate, domain-grounded technical instructions.

\begin{table}
\centering
\caption{Baseline vs. fine-tuned performance on the INGENZI\_DatasetV1.}
\label{tab:results}
\begin{tabular}{lcc}
\textbf{Metric} & \textbf{Baseline} & \textbf{Fine-Tuned} \\
F1 Score     & 0.22 & \textbf{0.38} \\
ROUGE-1      & 0.37 & \textbf{0.57} \\
ROUGE-2      & 0.18 & \textbf{0.41} \\
ROUGE-L      & 0.26 & \textbf{0.48} \\
BERTScore F1 & 0.86 & \textbf{0.91} \\
\end{tabular}
\end{table}

\section{Discussion}

This work demonstrates that domain-specific fine-tuning substantially improves medical foundation models for multimodal imaging equipment maintenance support. While the base MedGemma model exhibited moderate semantic alignment, it struggled with technical precision. Fine-tuning on the INGENZI dataset yielded major gains across all metrics, notably doubling ROUGE-2 (0.18 to 0.41), suggesting the model’s enhanced capacity to reproduce structured technical instructions. In general, the results show that a high-quality, targeted domain dataset such as INGENZI\_Dataset\_V1 can effectively specialize medical language models for complex operational tasks. This is critical for low-resource settings where original equipment manufacturer (OEM) support is limited. A lightweight model could serve as a vital decision-support tool for frontline technicians. Crucially, these improvements were achieved without retrieval augmentation (RAG), establishing a robust baseline for future hybrid systems.

\section{Limitations}

Several limitations remain. First, while the INGENZI dataset expands on prior work \cite{ref_bernes_et_al} by incorporating both Siemens MRI and Philips ultrasound documentation, validation across other modalities (e.g., CT, X-ray) and OEMs is still required. Second, evaluation was restricted to the MedGemma baseline, and as such, comparative benchmarks against other open-source or commercial LLMs are needed. Third, the lack of a live RAG framework leaves a minor risk of hallucinated or outdated instructions during inference. Finally, model performance was assessed solely via automated metrics, and rigorous expert human validation by biomedical engineers remains a prerequisite for real-world deployment. Our ongoing work integrates a hybrid weighted ensemble system that pairs this fine-tuned model with a RAG pipeline to enforce strict factual grounding.  Future work will scale the corpus across more modalities, integrate live retrieval grounding, and conduct field safety evaluations for deployment.

%
%
%
%

\end{document}